\documentclass[10pt,twocolumn,letterpaper]{article}

\usepackage{cvpr}
\usepackage{times}
\usepackage{epsfig}
\usepackage{graphicx}
\usepackage{amsmath}
\usepackage{amssymb}
\usepackage{amsthm}
\usepackage{amsfonts}
\usepackage{mathtools}
\usepackage{tabularx}
\usepackage{multirow}
\usepackage{bm}
\usepackage{arydshln}
\usepackage[hyphens]{url}
\usepackage{subcaption}

\graphicspath{{./images/}{./} }

\usepackage[breaklinks=true,colorlinks,bookmarks=false]{hyperref}

\cvprfinalcopy % *** Uncomment this line for the final submission

\DeclareMathOperator*{\argmin}{argmin}

\def\cvprPaperID{5180} % *** Enter the CVPR Paper ID here

\ifcvprfinal\fi
\begin{document}

%%%%%%%%% TITLE

%%%%%%%%% TITLE
\title{Generating Multiple Hypotheses for 3D Human Pose Estimation\\ with Mixture Density Network}

\author{Chen Li \quad \quad Gim Hee Lee\\
Department of Computer Science, National University of Singapore\\
{\tt\small \{lic, gimhee.lee\}@comp.nus.edu.sg}
}

\maketitle
%\thispagestyle{empty}

%%%%%%%%% ABSTRACT
\begin{abstract}
  3D human pose estimation from a monocular image or 2D joints is an ill-posed problem because of depth ambiguity and occluded joints. We argue that 3D human pose estimation from a monocular input is an inverse problem where multiple feasible solutions can exist. In this paper, we propose a novel approach to generate multiple feasible hypotheses of the 3D pose from 2D joints.
  In contrast to existing deep learning approaches which minimize a mean square error based on an unimodal Gaussian distribution, our method is able to generate multiple feasible hypotheses of 3D pose based on a multimodal mixture density networks. %Given 2D joint detection, our model is capable of generating multiple valid 3D human pose hypotheses. 
  Our experiments show that the 3D poses estimated by our approach from an input of 2D joints are consistent in 2D reprojections, which supports our argument that multiple solutions exist for the 2D-to-3D inverse problem. 
  Furthermore, we show state-of-the-art performance on the Human3.6M dataset in both best hypothesis and multi-view settings, and we demonstrate the generalization capacity of our model by testing on the MPII and MPI-INF-3DHP datasets. Our code is available at the project website\footnote{\url{https://github.com/chaneyddtt/Generating-Multiple-Hypotheses-for-3D-Human-Pose-Estimation-with-Mixture-Density-Network}}.
  
\end{abstract}

%%%%%%%%% BODY TEXT
\section{Introduction}

3D human pose estimation from a single RGB  image is an extensively studied problem in computer vision because of many potential useful real world applications such as forensic science, sports analysis and surveillance \etc. Significant progress in 3D human pose estimation has been made with deep learning  in the recent years.  One of the commonly used and effective deep learning based methods for 3D human pose estimation is the two-stage approach, where the 2D joints are first detected from the image input \cite{newell2016stacked, wei2016convolutional} followed by the 3D joint estimations from the detected 2D joints \cite{akhter2015pose,zhou2016sparseness, bogo2016keep,martinez2017simple,hossain2018exploiting, chen2016synthesizing,yasin2016dual, moreno20173d}. The advantage of the two-stage approach is that it decouples the harder problem of 3D depth estimation from the easier 2D pose estimation. In particular, variations in background scene, lighting, clothing shape, skin color \etc are removed before the 3D joint estimation stage. Furthermore, the model can be trained on different domains, \eg indoor and outdoor, with 2D annotations that are readily available.

\begin{figure}[t]
\begin{center}
  \includegraphics[width=1.0\linewidth]{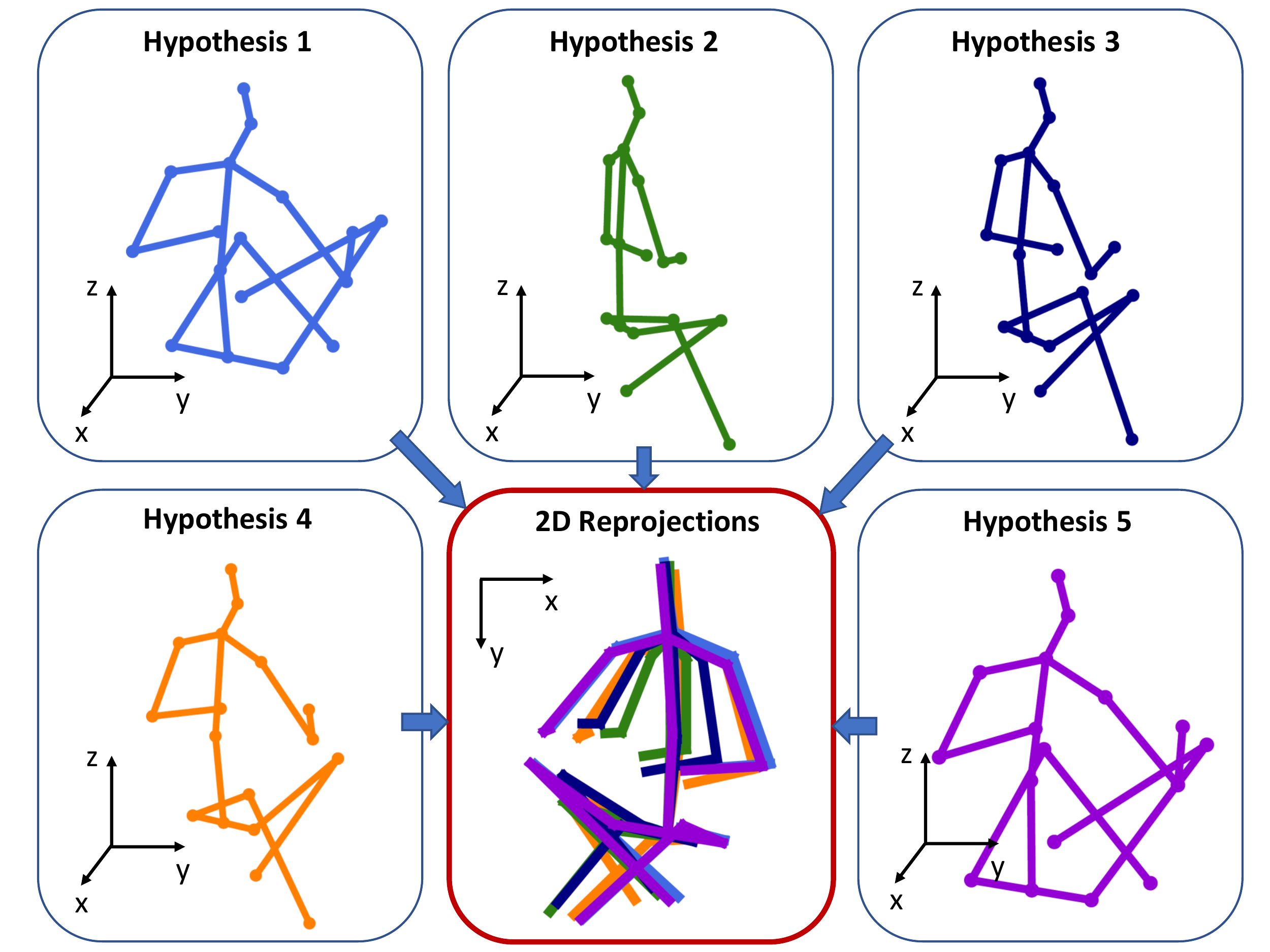}
\end{center}
\vspace{-3mm}
   \caption{An example of multiple feasible 3D pose hypotheses generated from our network reprojecting into similar 2D joint locations. (Best view in color)}

\label{fig:onecol}
\vspace{-4mm}
\end{figure}

Despite the significant progress with deep learning, 3D human pose estimation remains as a very challenging task due to the ambiguity in recovering 3D information from a single RGB image. More specifically, recovering 3D information from a single RGB image or 2D joint locations is an inverse problem \cite{bishop1994mixture} where multiple solutions may exist for the depth of a 3D joint along the light ray that reprojects onto the same 2D joint location, as illustrated in Figure~\ref{fig:onecol}.   The problem is further aggravated by the non-rigidity of the human pose and joint occlusions on the 2D image. Consequently, there could be many solutions of the 3D pose that satisfy the same 2D pose on an image, even after eliminating the infeasible 3D pose solutions by enforcing various geometric constraints, \eg joint limits \cite{akhter2015pose} and bone ratio \cite{zhou2017towards} \etc. In view of the inherent ambiguity of the 3D human pose estimation problem, we argue that it is more reasonable to design a model that generates multiple hypotheses of geometrically feasible 3D human pose that are consistent with the detected 2D joints from a single RGB image. In contrast, the widely adopted single estimate for the inverse problem with inherent ambiguity could lead to overfitting the model to the training data, and might not generalize well. This idea of generating multiple 3D pose hypotheses was first suggested very recently by Jahangiri and Yuille in \cite{jahangiri2017generating}.    

To this end, we introduce the mixture density networks (MDN) \cite{bishop1994mixture, ye2017occlusion} to the 3D joint estimation module of the two-stage approach. Contrary to most existing works that generate a single 3D pose by minimizing the negative log-likelihood of an unimodal Gaussian, \ie a mean squared error, we propose to estimate multiple hypotheses of the 3D pose by minimizing the negative log-likelihood of a multimodal mixture-of-Gaussians. The outputs of our mixture model is a set of mixing coefficients and parameters of the Gaussian kernels, \ie means and variances. The set of 3D pose hypotheses are given by the means of the Gaussian kernels, and the mixing coefficient and variances represent the uncertainties of each 3D pose hypothesis. 
% Specifically, we first use the ``Stacked Hourglass'' \cite{newell2016stacked} network to estimate the 2D joints from the input RGB image. Next, we use our mixture density network to generate multiple 3D pose hypotheses from the estimated 2D joints. 
Specifically our network consists of a feature extractor to lift the 2D joints into a feature space,  and a hypotheses generator to generate multiple hypotheses.  
The whole network is a simple network made up of several %
%layers of multi-layer perceptrons (mlps), 
linear layers with different non-linear activation units.

We show that our network achieves state-of-the-art results on the Human3.6M dataset \cite{ionescu2014human3} in both best hypothesis and multi-view settings. We also report results of our network on the outdoor MPII \cite{andriluka20142d} dataset and the MPI-INF-3DHP \cite{mehta2017monocular} dataset, where 3D pose labels are not used for training the network. Furthermore, we show the robustness of our network by applying it to scenarios where one or two limb joints are occluded/missing. Our main contributions are as follows: 

\begin{itemize}
    \item We explore the idea of generating multiple 3D pose hypotheses to alleviate the ambiguity problem that has not received much attention in the literature. 
    \item To the best of our knowledge, we are the first to introduce the mixture density model into 3D human pose estimation, which is more powerful than the single Gaussian distribution. 
    \item Our network achieves state-of-the art results on Human3.6M dataset in both best hypothesis and multi-view settings, and in cases where one or two limb joints are occluded/missing. 
\end{itemize}

\begin{figure*}
\begin{center}
%\fbox{\rule{0pt}{2in} \rule{.9\linewidth}{0pt}}
\includegraphics[width=0.98\linewidth]{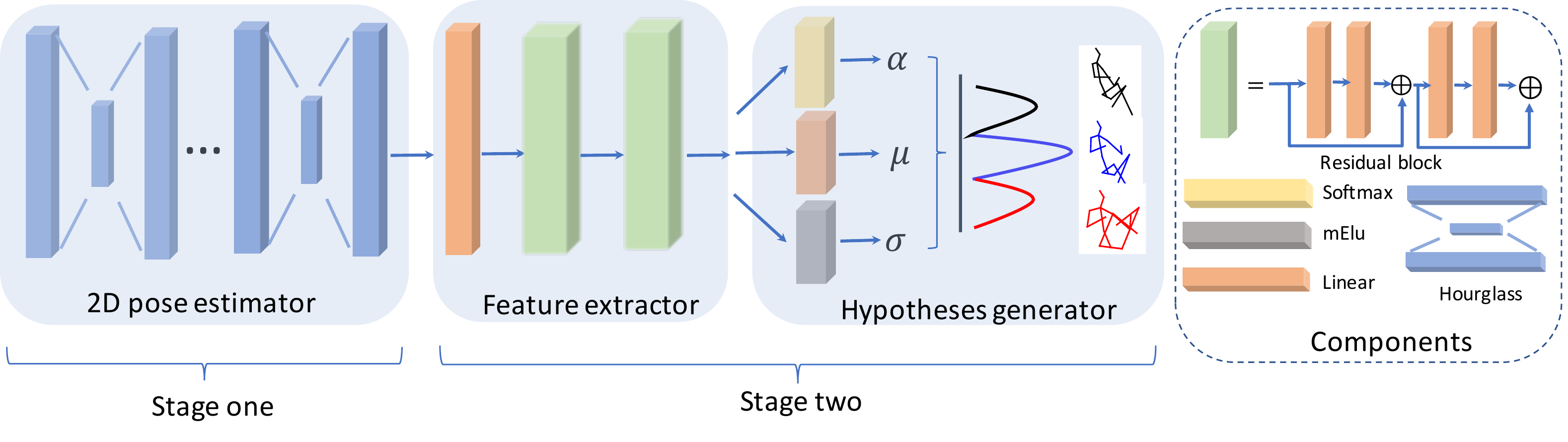}

\end{center}
\vspace{-6mm}
   \caption{Our network consists of a feature extractor and a 3D pose hypotheses generator, it generates multiple pose hypotheses from the 2D joints detected by 2D pose estimator.}
\label{fig:OurNetwork}
\vspace{-4mm}
\end{figure*}

%------------------------------------------------------------------------
\section{Related Work}
Existing human 3D pose estimation approaches fall into two categories according to their training techniques. The first category is to train deep convolutional neural networks (CNNs) end-to-end to estimate 3D human poses directly from the input images~\cite{pavlakos2017coarse,mehta2017monocular,zhou2016deep,park20163D, zhou2017towards, lee2018propagating, sun2018integral}. Zhou \etal.~\cite{zhou2016deep} use a sparse representation for 3D poses and predict the 3D pose with an expectation-maximization (EM) algorithm. The 2D poses are regarded as a hidden variable in the EM algorithm to remove the need of synchronized 2D-3D data. Park \etal~\cite{park20163D} improve conventional CNNs by concatenating 2D pose estimation as well as information on relative positions with respect to multiple joints. Pavlakos \etal~\cite{pavlakos2017coarse} use volumetric representation to represent 3D poses and adopt the stacked hourglass network~\cite{newell2016stacked}, which is originally designed for 2D pose estimation, to predict 3D volumetric heatmaps. Mehta \etal~\cite{mehta2017monocular} use transfer learning to transfer the knowledge learned for 2D pose estimation to the task of 3D pose estimation. Similarly, Zhou \etal~\cite{zhou2017towards} propose a weakly-supervised transfer learning method that uses mixed 2D and 3D labels. The 2D pose estimation sub-network and 3D depth regression sub-network share the same features such that the 3D pose labels for indoor environments can be transferred to in-the-wild images. The direct approach benefits from the rich information contained in images, \eg the front-back orientation of limbs. However, it will also be affected by a number of factors such as background, lighting, clothing \etc. A network trained on one dataset can not be generalized well to the other datasets with different environment, for example from indoor and outdoor environment. 

The second category \cite{akhter2015pose,zhou2016sparseness, bogo2016keep,martinez2017simple,hossain2018exploiting, chen2016synthesizing,yasin2016dual, moreno20173d} decouples 3D pose estimation into the well-studied 2D joint detection \cite{newell2016stacked, wei2016convolutional} and 3D pose estimation from the detected 2D joints. Akhter \etal \cite{akhter2015pose} propose a multi-stage approach to estimate the 3D pose from 2D joints using an over-complete dictionary of poses. Bogo \etal \cite{bogo2016keep} estimate 3D pose by first fitting a statistical body shape model to the 2D joints, and then minimizing the error between the reprojected 3D model and detected 2D joints. Chen \cite{chen2016synthesizing} and Yasin \cite{yasin2016dual} regard 3D pose estimation as a matching between the estimated 2D pose and the 3D pose from a large pose library. Martinez \etal\cite{martinez2017simple} design a simple fully connected residual network to regress 3D pose from 2D joint detections. The decoupled approach can make use of both indoor and in-the-wild images to train the 2D pose estimators. More importantly, this approach is domain invariant since the input of the second stage is the 2D joints. However, estimating 3D pose from 2D joints is more challenging because 2D pose data contains less information than images, thus there are more ambiguities. 

% Moreover, 3D pose estimation from 2D joints itself is an inverse problem because multiple feasible 3D poses can have similar 2D reprojections. 

To solve the ill-posed problem of estimating 3D pose from 2D joints, Jahangiri and Yuille \cite{jahangiri2017generating} first proposed to generate multiple diverse pose hypotheses. They first learned a 3D Gaussian mixture model (GMM) model~\cite{sigal2004tracking} from a uniformly sampled set of Human3.6M poses, and then use conditional sampling to get samples of the 3D poses with reprojected joints errors that are within a threshold. Inspired by their work, we solve the ambiguity problem by generating multiple hypotheses. Instead of using the traditional GMM approach, we introduce the MDN which was first proposed in \cite{bishop1994mixture}. The MDN can represent arbitrary conditional distributions by combining a conventional neural network with a mixture density model. Ye \etal \cite{ye2017occlusion} used a hierarchical MDN to solve the occlusion problem in hand pose estimation. Inspired by the work of Ye \etal, we use the MDN to solve the depth ambiguity and occlusion problem in 3D human pose estimation. 
\vspace{-1mm}
\section{Our Mixture Density Network}
Figure~\ref{fig:OurNetwork} shows the illustration of our deep network to generate multiple hypotheses for 3D human pose estimation. Our network follows the commonly used two-stage approach that first estimates the 2D joints from the input images followed by the 3D pose estimation from the estimated 2D joints. We adopt the state-of-the-art stacked hourglass \cite{newell2016stacked} network as the 2D joint estimation module, and use our MDN which consists of a feature extractor and a hypotheses generator to generate the multiple 3D pose hypotheses. Given the 2D joint detections $\textbf{x} \in \mathbb{R}^{2N}$, where $N$ is the number of joints in one pose, our goal is to learn a function $f:\textbf{x} \mapsto \Theta$ which maps $\textbf{x}$ into a set of output parameters $\Theta = \{\bm{\mu}, \bm{\sigma}, \bm{\alpha} \}$ for our mixture model. $\bm{\mu} = \{\mu_1, ..., \mu_M \mid \mu_i\in \mathbb{R}^{3N}\}$, $\bm{\sigma} = \{\sigma_1, ..., \sigma_M\ \mid \sigma_i \in \mathbb{R}\}$ and $\bm{\alpha} = \{\alpha_1, ..., \alpha_M\ \mid 0 \leq \alpha_i \leq 1,~\sum_i \alpha_i = 1\}$ are the means, variances and mixing coefficients of the mixture model. $M$ is the number of Gaussian kernels.   
The mean of each Guassian kernel $\mu_i \in \bm{\mu}$ represents one 3D pose hypothesis, and the number of Gaussian kennels $M$ decides the number of hypotheses generated by our model.

\subsection{Model Representation}
The probability density of the 3D pose $\textbf{y} \in \mathbb{R}^{3N}$ given the 2D joints $\textbf{x} \in \mathbb{R}^{2N}$ is represented as a linear combination of Gaussian kernel functions
\begin{align}
  \label{eq:prob_density}
  p(\textbf{y} \mid \textbf{x}) = \sum_{i=1}^{M}\alpha_i(\textbf{x})\phi_i(\textbf{y} \mid \textbf{x}),
\end{align}
where $M$ is the number of Gaussian kernels, \ie the number of hypotheses. $\alpha_i(\textbf{x})$ is the mixing coefficients, which can be regarded as a prior probability of a 3D pose data $\textbf{y}$ being generated from the $i^{th}$ Gaussian kernel given the input 2D joints $\textbf{x}$. Here $\alpha_i(\textbf{x})$ must satisfy the constraint
\begin{align}
  \label{eq:alpha}
  0\leq\alpha_i(\textbf{x})\leq 1,~~~~\sum_{i=1}^{M}\alpha_i(\textbf{x}) = 1.
\end{align}
$\phi_i(\textbf{y} \mid \textbf{x})$ is the conditional density of the 3D pose $\textbf{y}$ for the $i^{th}$ kernel, which can be expressed as a Gaussian distribution
\begin{align}
  \label{eq:gaussion_kernel}
  \phi_i(\textbf{y} \mid \textbf{x}) = \frac{1}{(2\pi)^{d/2}\sigma_i(\textbf{x})^d}\exp{-\frac{\|{\textbf{y}-\mu_i(\textbf{x})}\|^2}{2\sigma_i(\textbf{x})^2}}.
\end{align}
$\mu_i(\textbf{x})$ and $\sigma_i(\textbf{x})$ denote the mean and variance of the $i^{th}$ kernel, respectively. $d$ is the dimension of the output 3D pose $\textbf{y}$. All the parameters of the mixture model, including the mixing coefficients $\alpha_i(\textbf{x})$, the mean $\mu_i(\textbf{x})$ and the variance $\sigma_i(\textbf{x})$ are functions of the input 2D pose $\textbf{x}$. 

Note that the mixture model degenerates to a single Gaussion distribution when the means and variances of 
all Gaussian kernels are similar, \ie $\mu_i(\textbf{x}) \approx \mu(\textbf{x})$, %$\sigma(\textbf{x}) \approx \sigma_i(\textbf{x})$ 
$\sigma_i(\textbf{x}) \approx \sigma(\textbf{x})$ 
%and $\alpha_i(\textbf{x}) \approx \frac{1}{M}$
for $i = 1, ..., M$. Hence, 
\begin{align}
  \label{eq:prob_density_single}
  p(\textbf{y} \mid \textbf{x}) & = \sum_{i=1}^{M}\alpha_i(\textbf{x})\phi_i(\textbf{y} \mid \textbf{x}) \notag\\
  & \approx \sum_{i=1}^{M}\alpha_i(\textbf{x}) \mathcal{N}(\mu(\textbf{x}), \sigma(\textbf{x})) \\
 & = \mathcal{N}(\mu(\textbf{x}), \sigma(\textbf{x})) \notag.
\end{align}
Specifically in our case, the 3D pose hypotheses generated by the MDN will collapse into approximately a single Gaussian when the given 2D pose is simple and less ambiguous, \eg no occlusions and/or missing joints.  

\subsection{Network Architecture}
From Eqn.~\eqref{eq:prob_density}, \eqref{eq:alpha} and \eqref{eq:gaussion_kernel}, we can see that all parameters $\Theta(\textbf{x})=\{\bm{\mu}(\textbf{x}), \bm{\sigma}(\textbf{x}), \bm{\alpha}(\textbf{x}) \}$ of the Gaussian mixture distribution of $\textbf{y}$ are functional form of $\textbf{x}$. Hence, we learn this function $f:\textbf{x}\mapsto\Theta$ using a deep network which can be expressed as
\begin{align}
  \label{eq:mapping_func}
  \Theta =  f(\textbf{x};\textbf{w}),
\end{align}
\noindent where $\textbf{w}$ is the set of learnable weights in the deep network. The 
probability density in Eqn. \eqref{eq:prob_density} can be rewritten to include the learnable weights $\textbf{w}$ of the deep network, \ie, 
\begin{equation}
\label{eq:probDensityDeep}
    p(\textbf{y} \mid \textbf{x}, \textbf{w}) = \sum_{i=1}^{M}\alpha_i(\textbf{x}, \textbf{w})\phi_i(\textbf{y} \mid \textbf{x}, \textbf{w}),
\end{equation}
\noindent where
\begin{equation}
\phi_i(\textbf{y} \mid \textbf{x}, \textbf{w}) = \frac{1}{(2\pi)^{d/2}\sigma_i(\textbf{x}, \textbf{w})^d}\exp{-\frac{\|{\textbf{y}-\mu_i(\textbf{x}, \textbf{w})}\|^2}{2\sigma_i(\textbf{x}, \textbf{w})^2}}.
\end{equation}

\noindent The parameters $\Theta(\textbf{x}, \textbf{w}) = \{\bm{\mu}(\textbf{x}, \textbf{w}), \bm{\sigma}(\textbf{x}, \textbf{w}), \bm{\alpha}(\textbf{x}, \textbf{w}) \}$ are now dependent on the learnable weights $\textbf{w}$ of the deep network $f(\textbf{x}; \textbf{w})$.

We modify the 3D pose estimation module in \cite{martinez2017simple} to form our deep network $f(\textbf{x};\textbf{w})$.
%use the conventional 2D-to-3D pose estimator\cite{martinez2017simple} to learn the features of 2D inputs, the outputs are $\{\alpha_i(x), \mu_i(x), \sigma_i(x)\}$ for each component of the mixture model. Similar to \cite{martinez2017simple},
More specifically, our approach is a simple multilayer neural network. Given an input of 2D joints $\textbf{x} \in \mathbb{R}^{2N}$, we use one linear layer to map the input into an 1024 dimensional feature space, followed by two residual blocks which respectively consists of a linear layer, batch normalization , dropout , and Rectified Linear Units. And there are residual connections between the input and output of each residual block. Different from \cite{martinez2017simple} which adds another linear layer to directly regress the 3D pose $\textbf{y} \in \mathbb{R}^{3N}$ from the feature space, our network estimates the parameters $\Theta$ of the mixture model. In particular, we use different activation functions to satisfy the constraints of the three parameters  $\Theta(\textbf{x}, \textbf{w}) = \{\bm{\mu}(\textbf{x}, \textbf{w}), \bm{\sigma}(\textbf{x}, \textbf{w}), \bm{\alpha}(\textbf{x}, \textbf{w}) \}$. Specifically, we use a normal linear layer for parameter $\bm{\mu}(\textbf{x}, \textbf{w})$, a softmax function for the mixture coefficient $\bm{\alpha}(\textbf{x}, \textbf{w})$ so that it lies in the range of ${[0, 1]}$ and sums up to $1$, and a modified ELU function \cite{clevert2015fast} defined as:

\begin{equation}
  \label{eq:Elu_func}
   h(t) =
  \begin{cases}
     t+1, & \text{if } t\geq 0\\
    \gamma(\exp(t) - 1) +1 , & \text{otherwise}
  \end{cases}
\end{equation}
for the variance $\bm{\sigma}(\textbf{x}, \textbf{w})$ to keep it positive. Here, $\gamma$ is a scale for negative factor.

\subsection{Optimization}
 Given a training dataset with $K$ pairs of ground truth labels for the corresponding 2D joints $\textbf{X}$ and 3D poses $\textbf{Y}$, \ie $\{\textbf{X}, \textbf{Y}\} = \{\{\textbf{x}_j, \textbf{y}_j\} \mid j = 1, ..., K\}$, the objective is to find the maximum a posterior of the set of learnable weights $\textbf{w}$. More formally, assuming that each training data is independent and identically distributed (i.i.d), the posterior distribution of $\textbf{w}$ is given by
 \begin{align}
     p(\textbf{w} \mid \textbf{X}, \textbf{Y}, \Psi) &\propto p(\textbf{Y} \mid \textbf{X}, \textbf{w})p(\textbf{w} \mid \textbf{X}, \Psi) \\ \notag
     &= p(\textbf{w} \mid \textbf{X}, \Psi)\prod_{j=1}^{K}p(\textbf{y}_j \mid \textbf{x}_j, \textbf{w}) \\ \notag
     &= p(\textbf{w} \mid \textbf{X}, \Psi) \prod_{j=1}^{K}\sum_{i=1}^{M}\alpha_i(\textbf{x}_j, \textbf{w})\phi_i(\textbf{y}_j \mid \textbf{x}_j, \textbf{w}),
 \end{align}
 
 \noindent where $\Psi$ is the hyperparameter of the prior over the learnable weights $\textbf{w}$. Hence, the optimal weight $\textbf{w}^*$ can be obtained from the minimization of the negative log-posterior
 \begin{align}
    \textbf{w}^* &= \argmin_{\textbf{w}} -\ln{p(\textbf{w} \mid \textbf{X}, \textbf{Y}, \Psi)} \\ \notag
    &= \argmin_{\textbf{w}} \underbrace{-\sum_{j=1}^{K}\ln{p(\textbf{y}_j \mid \textbf{x}_j, \textbf{w})} - \ln{p(\textbf{w} \mid \textbf{X}, \Psi)}}_{\mathcal{L}},
 \end{align}
 
 \noindent where $\mathcal{L}$ is taken to be the loss function for training our deep network $f(\textbf{x}; \textbf{w})$. More specifically, 
 \begin{align}
    \mathcal{L} &= -\sum_{j=1}^{K}\ln{p(\textbf{y}_j \mid \textbf{x}_j, \textbf{w})} - \ln{p(\textbf{w} \mid \textbf{X}, \Psi)} \\ \notag
    &= -\sum_{j=1}^{K}\ln{\sum_{i=1}^{M}\alpha_i(\textbf{x}_j, \textbf{w})\phi_i(\textbf{y}_j \mid \textbf{x}_j, \textbf{w})} - \ln{p(\textbf{w} \mid \textbf{X}, \Psi)} \\ \notag
    &= \mathcal{L}_{\text{3D}} + \mathcal{L}_{\text{prior}}.
 \end{align}
 \noindent The prior loss $\mathcal{L}_{\text{prior}}$ can be further evaluated into:
 \begin{align}
     \mathcal{L}_{\text{prior}} &= - \ln{p(\textbf{w} \mid \textbf{X}, \Psi)} \\ \notag
     &= -\ln{p(\textbf{w}, \textbf{X} \mid \Psi) } + \ln{p(\textbf{X} \mid \Psi)} \\ \notag 
     &\propto -\ln{p(\Theta(\textbf{w}, \textbf{X}) \mid \Psi) } \\ \notag
     &= -\ln{p(\bm{\alpha}(\textbf{w}, \textbf{X}) \mid \Psi)} - \ln{p(\bm{\mu}(\textbf{w}, \textbf{X}), \bm{\sigma}(\textbf{w}, \textbf{X}) \mid \Psi)},
 \end{align}
 \noindent where the term $\ln{p(\textbf{X} \mid \Psi)}$ can be dropped in the loss function since it is independent of $\textbf{w}$, and we write the random variables $\{\textbf{w}, \textbf{X}\}$ in its functional form $\Theta(\textbf{w}, \textbf{X})$ given by the deep network. We further assume a uniform prior over $\bm{\mu}(\textbf{w}, \textbf{X})$ and $\bm{\sigma}(\textbf{w}, \textbf{X})$, and a Dirichlet conjugate prior over the mixing coefficients $\bm{\alpha}(\textbf{w}, \textbf{X})$ that follows a Categorical distribution, we get 
\begin{align}
     \mathcal{L}_{\text{prior}} &= -\ln{p(\bm{\alpha}(\textbf{w}, \textbf{X}) \mid \Lambda)} \\ \notag
     &= - \sum_{j=1}^{K}\ln{p(\alpha_{1}(\textbf{w}, \textbf{x}_j), ..., \alpha_{M}(\textbf{w}, \textbf{x}_j) \mid \Lambda)}, 
\end{align}
\noindent where
\begin{align}
  \label{eq:dirichlet}
p(\alpha_{1}, ..., \alpha_{M} \mid \Lambda) &= \text{Dir}_{[\alpha_1, ..., \alpha_M]}{[\lambda_1,...,\lambda_M]} \\ \notag
&= \frac{\Gamma{[\sum_{i=1}^{M}\lambda_i]}}{\prod_{i=1}^{M}\Gamma{[\lambda_i]}}\prod_{i=1}^{M}\alpha_i(\textbf{w}, \textbf{x}_j)^{\lambda_i -1}.
\end{align}
 \noindent $\Gamma{[.]}$ is the Gamma function, and $\Lambda = \{\lambda_1, ..., \lambda_M\}$ are the hyperparameters of the Dirichlet distribution, where $\lambda_i >0$ for $1\leq i \leq M$. The total loss function to train our deep network is given by
 \begin{align*}
     \mathcal{L} &= \mathcal{L}_{\text{3D}} + \mathcal{L}_{\text{prior}},
 \end{align*}
 \noindent where
\begin{align}
 &\mathcal{L}_{\text{3D}}
     = -\sum_{j=1}^{K}\ln{\sum_{i=1}^{M}\alpha_i(\textbf{x}_j, \textbf{w})\phi_i(\textbf{y}_j \mid \textbf{x}_j, \textbf{w})} \\ \notag
&\mathcal{L}_{\text{prior}} =     
     - \sum_{j=1}^{K}\sum_{i=1}^M(\lambda_i - 1)\ln{\alpha_i(\textbf{w}, \textbf{x}_j)}.
\end{align}
Note that we drop $\frac{\Gamma{[\sum_{i=1}^{M}\lambda_i]}}{\prod_{i=1}^{M}\Gamma{[\lambda_i]}}$ in $\mathcal{L}_{\text{prior}}$ because it is independent of $\textbf{w}$.

\paragraph{Remarks:} The term $\mathcal{L}_{\text{prior}}$ regularizes the mixing coefficients of our mixture model. Setting $\lambda_i = 1$ for $i=1, .., M$ implies that we have no prior knowledge over the mixing coefficients. In our experiments, we set $\lambda_1 = ... = \lambda_M = C$, where $C>1$ is a constant scalar value to prevent overfitting of a single Gaussian kernel in the MDN to the training data, \ie a single mixing coefficient $\alpha_i \approx 1$ and $\alpha_{j \neq i} \approx 0$.

% It should be noted that it is possible to include the training loss for the 2D joint estimation module $\mathcal{L}_{2D}$ from \cite{newell2016stacked} (albeit optional and not necessarily better results) into the total loss to train our network end-to-end with the input images $\textbf{I}=\{I_1, ..., I_K\}$, 2D joints $\textbf{X}$ and 3D poses $\textbf{Y}$.
% Hence, the total training loss becomes 
% \begin{align}
%      \mathcal{L} &= \beta\mathcal{L}_{2D} + \mathcal{L}_{3D} + \mathcal{L}_{prior},
%  \end{align}
% \noindent where $\beta$ is a scaling factor for the 2D loss that is in the image pixel space. The soft argmax function can be used to get the 2D joints from heatmaps in this case.

%We did not give any weight to balance the two losses. The reason is that the weight of the prior loss $L_{\text{prior}}$ has already implicitly included in the hyperparameters$\{\lambda_1,..., \lambda_m\}$. Thus we don't include another weight for $L_{\text{prior}}$. We give a more detailed description about it in our supplementary material.

\begin{table*}
\vspace{-2mm}
\caption{Quantitative results of MPJPE in millimeter on Human3.6M under protocol \# 1 and \# 2. (Best result in bold)}
\vspace{-2.5mm}
\centering
\small
\setlength{\tabcolsep}{1.35pt}
\begin{tabular*}{0.99\textwidth}{ l c c c c c c c c c c c c c c c c } 
 
 \hline
 Protocol \#1 &  Direct. & Discuss & Eating & Greet & Phone & Photo & Pose & Purch. &  Sitting & SittingD. & Smoke & Wait & WalkD. & Walk & WalkT. & Avg.\\ 
 \hline
 LinKDE \etal\cite{ionescu2014human3} &132.7 & 183.6 & 132.3 & 164.4 & 162.1 & 205.9 & 150.6 & 171.3 & 151.6 & 243.0 & 162.1 & 170.7 & 177.1 & 96.6 & 127.9 & 162.1 \\
 Du \etal\cite{du2016marker} & 85.1 & 112.7 & 104.9 &122.1 & 139.1 & 135.9 &105.9 & 166.2 & 117.5 & 226.9 & 120.0 & 117.7 & 137.4 & 99.3 & 106.5 & 126.5 \\
 Zhou \etal\cite{zhou2016deep} & 87.4 & 109.3 & 87.1 & 103.2 & 116.2 & 143.3 & 106.9 & 99.8 & 124.5 & 199.2 & 107.4 & 118.1 & 114.2 & 79.4 & 97.7 & 113.0\\
 Pavlakos \etal\cite{pavlakos2017coarse} &67.4 & 71.9 & 66.7 & 69.1 & 72.0 & 77.0 & 65.0 & 68.3 & 83.7 & 96.5 & 71.7 & 65.8 & 74.9 & 59.1 & 63.2 & 71.9 \\
 Jahangiri \etal \cite{jahangiri2017generating} & 63.1 & 55.9 & 58.1 & 64.5 & 68.7 & 61.3 & 55.6 &86.1 & 117.6 &71.0 &71.2 & 66.3 & 57.1 & 62.5 & 61.0 & 68.0\\
 Zhou \etal \cite{zhou2017towards} & 54.8 & 60.7 & 58.2 & 71.4 & 62.0 & 65.5 & 53.8 & 55.6 & 75.2 & 111.6 & 64.1 & 66.0 & 51.4 & 63.2 & 55.3 & 64.9 \\
 Martinez \etal\cite{martinez2017simple} & 51.8 & 56.2 & 58.1 & 59.0 & 69.5 & 78.4 & 55.2 & 58.1 & 74.0 & 94.6 & 62.3 & 59.1 & 65.1 & 49.5 & 52.4 & 62.9 \\

 Lee \etal\cite{lee2018propagating} & \bf 43.8 & 51.7 & \bf 48.8 & 53.1 & \bf 52.2 & 74.9 & 52.7 & \bf 44.6 & \bf 56.9 & 74.3 & 56.7 & 66.4 & 47.5 & 68.4 & 45.6 & 55.8 \\
 Ours & \bf 43.8 & \bf 48.6 & 49.1  & \bf 49.8 & 57.6 & \bf 61.5  & \bf 45.9 & 48.3 & 62.0 & \bf 73.4 & \bf 54.8 & \bf 50.6 &  56.0 & \bf 43.4 & \bf 45.5 & \bf 52.7 \\\hline
 
Protocol \#2 &  Direct. & Discuss & Eating & Greet & Phone & Photo & Pose & Purch. &  Sitting & SittingD. & Smoke  & Wait & WalkD. & Walk & WalkT. & Avg.\\ 
 \hline
 Yasin \etal\cite{yasin2016dual} & 88.4  & 72.5  & 108.5 & 110.2 & 97.1 & 142.5& 81.6 & 107.2 & 119.0 & 170.8 & 108.2 &86.9 &92.1 &  165.7 &  102.0 &110.1\\
 Bogo \etal \cite{bogo2016keep} & 62.0 & 60.2 & 67.8 & 76.5 & 92.1 & 77.0 & 73.0 & 75.3 & 100.3 & 137.3 & 83.4 & 77.3 & 86.8 & 79.7 & 87.7 & 82.3\\
 Moreno \etal\cite{moreno20173d} & 66.1 &  61.7 & 84.5 & 73.7 & 65.2 & 67.2 & 60.9 & 67.3 & 103.5 & 74.6 & 92.6 & 69.6 & 71.5 & 78.0 & 73.2 & 74.0 \\
 
 Martinez \etal\cite{martinez2017simple} &39.5 &   43.2 & 46.4 & 47.0 &  51.0 & 56.0  & 41.4 & 40.6  & 56.5  &  69.4  & 49.2 &  45.0 &  49.5 & 38.0 &  43.1 & 47.7\\
 Lee \etal\cite{lee2018propagating} & 37.4  & \bf38.9  & 45.6 & 43.8 & 48.5 &  54.6  & 39.9 & 39.2 &  53.0 & 68.5 & 51.5 & \bf 38.4 & \bf 33.2 & 55.8 &  37.8 &45.7 \\
 
 Ours  & \bf 35.5 & 39.8  & \bf41.3  & \bf 42.3  & \bf 46.0  & \bf 48.9 & \bf 36.9  & \bf37.3 &  \bf 51.0
 & \bf 60.6 & \bf 44.9  &  40.2  &  44.1  & \bf 33.1  & \bf 36.9  & \bf 42.6\\
 \hline
 \vspace{-3mm}
\end{tabular*}
\label{Tab:MPJPE_Results_Human3.6}
\end{table*}

\section{Experiments}
Our model is implemented in Tensorflow, and we use the ADAM \cite{kingma2014adam} optimizer
with an initial learning rate of 0.001 and exponential decay. The batch size is set to 64 and we initialize the weights of linear layers with the Kaiming initialization~\cite{He_2015_ICCV}. The number of Gaussian kernels is set to 5 and the hyperparameters $\{\lambda_1, ... , \lambda_M\}$ in Eqn.~\eqref{eq:dirichlet} are set to 2. We train our network for 200 epoches with a dropout rate of 0.5. We also apply max-norm constraint on the weight of each layer so that it is in range ${[0,1]}$. Moreover, we clip the value of $\alpha_i(x)$ to $[1\mathrm{e}{-8}, 1]$ and $\sigma_i(x)$ to $[1\mathrm{e}{-15}, 1\mathrm{e}{15}]$ to prevent the training loss from becoming NaN. We also use the \textit{log-sum-exp} trick as previous work~\cite{brando2017mixture} to avoid 
the underflow problem.

%We show our results in table  Following \cite{martinez2017simple}, we also test our model on %groundtruth 2D input

\subsection{Datasets and Protocols}
We show numerical results for the Human3.6M dataset~\cite{ionescu2014human3} and compare with other state-of-the-art approaches. We also apply our approach to other datasets including MPII~\cite{andriluka20142d} and MPI-INF-3DHP~\cite{mehta2017monocular} datasets to test the generalization capacity of our network.
\vspace{-3mm}
\paragraph{Human3.6M dataset: } This is currently the
largest available video pose dataset, which provides accurate 3D body
joint locations recorded by a Vicon motion capture system. There are 15 activity
scenarios in total such as ``Walking'', ``Eating'', ``Sitting'' and ``Discussion'', each action is performed by 7 professional actors. Accurate 2D joint locations , 3D pose annotations and camera parameters are provided. Following \cite{martinez2017simple}, we apply standard normalization to the 2D inputs and 3D outputs by subtracting the mean and dividing by the standard deviation of the training data. We also zero-center the 3D poses around the hip joint since we do not predict the global position of the 3D pose.
\vspace{-3mm}
\paragraph{MPII dataset:} This is a standard dataset for 2D human pose estimation, which contains 25K unconstrained images collected from YouTube videos. This is the most challenging in-the-wild dataset and we use it to test the generalization of our approach. We report qualitative result for this datset because 3D pose information is not provided.
\vspace{-3mm}
\paragraph{MPI-INF-3DHP dataset:} It is a newly released 3D human pose dataset which is captured by a Mocap system in both indoor and outdoor scenes. We only use the test split of this dataset that includes 2935 frames from six subjects performing seven actions.
\vspace{-3mm}
\paragraph{2D detections:} We use the state-of-the-art stacked hourglass network \cite{newell2016stacked} to get the 2D joint detections. The stacked hourglass network is pretrained on the MPII dataset and then fine-tuned on the Human3.6M dataset.
\vspace{-3mm}
\paragraph{Evaluation protocols:}  For the Human3.6M dataset, we follow the standard protocol of using S1, S5, S6, S7 and S8 for training, and S9 and S11 for testing. The evaluation metric is the Mean Per Joint Position Error (MPJPE) in millimeter between the ground truth and the estimated 3D pose. Since our network generating multiple hypotheses for each 2D detection, we follow \cite{jahangiri2017generating} to compute the MPJPE between the ground truth and the best 3D hypothesis generated by our network. The 3D Percentate of Correct Keypoints (3DPCK) \cite{mehta2017monocular} is adopted as the metric for the MPI-INF-3DHP dataset . 
% \vspace{-1mm}

% \begin{table*}
% \caption{Quantitative comparisons of MPJPE in millimeter on Human3.6M under Protocol \# 1 (best result in bold)}
% \centering
% \small
% \setlength{\tabcolsep}{1.9pt}
% \begin{tabular*}{0.98\textwidth}{ c c c c c c c c c c c c c c c c c } 
 
%  \hline
%  Methods &  Direct. & Discuss & Eating & Greet & Phone & Pose & Purch. &  Sitting & SittingD. & Smoke & Smoke & Wait & WalkD. & Walk & WalkT. & Avg.\\ 
%  \hline
 
%   Pavlakos et al.\cite{DBLP:journals/corr/PavlakosZDD17} & 41.2 & 49.2 & \bf 42.8  & \bf 43.4  &  55.6  & \bf 46.9 & \bf 40.3 & 63.7 & 97.6 & 119.9
%     & 52.1 & \bf 42.7 & 51.9 & 41.8 & \bf 39.4 & 56.9\\
%  Lee et al.\cite{lee2018propagating}  & 40.2 & 49.2 & 47.8 & 52.6 & \bf 50.1 & 75.0 & 50.2 & 43.0 & \bf 55.8 & 73.9 & 54.1 & 55.6 & \bf 43.3 & 58.2 & 43.3 & 52.8 \\
%  Hossain et al.\cite{hossain2018exploiting} & 44.2 &  46.7 & 52.3 & 49.3 & 59.9 & 59.4 & 47.5 & 46.2 & 59.9 & \bf 65.6 & 55.8 & 50.4  & 52.3 & 43.5 & 45.1 & 51.9 \\

%  Ours & \bf 39.9 & \bf 43.9 & 46.3 & 47.0 & 52.6 & 61.7 & 42.3 & \bf 46.8 & 58.7 & 74.1 & \bf 49.5 & 47.3 & 52.3 & \bf 39.5 & 41.3 & \bf 49.6 \\\hline
 
% \end{tabular*}
% \label{table3}
% \end{table*}

\begin{table}[htp]
\caption{Results by using multi-view information}
\vspace{-2mm}
\centering
\small
\setlength{\tabcolsep}{3.8pt}
\begin{tabular*}{0.45\textwidth}{ c c c c c } 
 \hline
% &\multicolumn{1}{c}{Methods}\vline
% &\multicolumn{3}{Temporal}{Methods}\vline
% &\multicolumn{2}{c}{Muti-view}\vline \\
 Methods  & Lee\cite{lee2018propagating} & Hossain\cite{hossain2018exploiting} &  Pavlakos \cite{pavlakos2017harvesting} & Ours\\ 
 \hline

  Avg. & 52.8 & 51.9 & 56.9  & \bf 49.6  \\
  \hline
  \vspace{-5mm}

\end{tabular*}
\label{Tab:multi-view}
\end{table}

\begin{table*}[h!]
\caption{Results with one (the first three rows) or two (the last three rows) missing joints}
\vspace{-1mm}
\centering
\small
\setlength{\tabcolsep}{1.35pt}
\begin{tabular*}{0.99\textwidth}{ l c c c c c c c c c c c c c c c c } 
 
 \hline
 &  Direct. & Discuss & Eating & Greet & Phone & Smoke & Pose & Purch. &  Sitting & SittingD.  & Smoke & Wait & WalkD. & Walk & WalkT. & Avg.\\ 
 \hline
 
 Jahangiri \etal \cite{jahangiri2017generating}  & 108.6 & 105.9 &105.6 &109.0 &105.5 & 109.9 &102.0 & 111.3 & 119.6 &107.8 & 107.1 &111.3 &108.4 & 107.0 & 110.3 & 108.6\\
 Martinez \etal\cite{martinez2017simple} & 57.4 & 61.6 & 64.3 & 65.6 & 73.3 &85.5 & 61.0 & 62.1 & 84.0 & 101.1 & 68.2 & 66.7 & 70.8 & 55.6 & 59.6 & 69.1\\
 Ours & \bf48.9 & \bf53.9 & \bf54.5 & \bf55.5 & \bf 62.6 & \bf 70.4 & \bf 51.3 & \bf 52.0 & \bf 69.7 & \bf 83.9 & \bf 60.7 & \bf 57.2 & \bf 62.4 & \bf 48.3 & \bf 50.8 & \bf 58.8 \\ \hdashline
 
 Jahangiri \etal \cite{jahangiri2017generating}  & 125.0 &121.8 &115.1 &124.1 &116.9 &123.8 &116.4 &119.6 &130.8 & 120.6 & 118.4 & 127.1 &125.9 &121.6 &127.6 & 122.3\\
 Martinez \etal\cite{martinez2017simple} 
 & 62.9 &66.9 & 69.9 &71.4 &80.2 & 93.8 &66.3 &65.9 &90.6 & 109.7 &74.2 &72.1 &75.5 &61.7 &65.7 &75.1 \\
 Ours & \bf 54.0 & \bf 58.5 & \bf 60.6 & \bf 61.4 & \bf 68.6 & \bf 77.9 & \bf 56.6 & \bf 57.0 & \bf 77.8 & \bf 92.4 & \bf 66.2 & \bf 62.6 &\bf 67.5 & \bf 52.5 & \bf 55.0 & \bf 64.6 \\ \hline

 \vspace{-3mm}
\end{tabular*}
\label{Tab:missingJoints}
\end{table*}

\begin{table}[h!]
\caption{Quantitative results on MPI-INF-3DHP dataset }
\vspace{-1mm}
\centering
\small
\setlength{\tabcolsep}{2pt}
\begin{tabular*}{0.45\textwidth}{ l c c c c } 
 \hline

   & Studio GS & Studio no GS & Outdoor & All PCK\\ 
 \hline
 Mehta \etal\cite{mehta2017monocular} & \bf 84.1 & \bf 68.9 & 59.6  &  \bf 72.5  \\
 Ours & 70.1 & 68.2 & \bf 66.6 & 67.9\\
  \hline
 \vspace{-3mm}
\end{tabular*}
\label{Tab:resultsMPI-INF-3DHP}
\end{table}

\begin{figure*}[h!]
  \centering
% \quad 
% \includegraphics[width=0.29\textwidth]{latex/images/sample_5-crop.pdf} \hfill
%   \includegraphics[width=0.29\textwidth]{latex/images/sample_8-crop.pdf}\hfill
% \includegraphics[width=0.29\textwidth]{latex/images/sample_112-crop.pdf}\quad\\

\quad 
\includegraphics[width=0.29\textwidth]{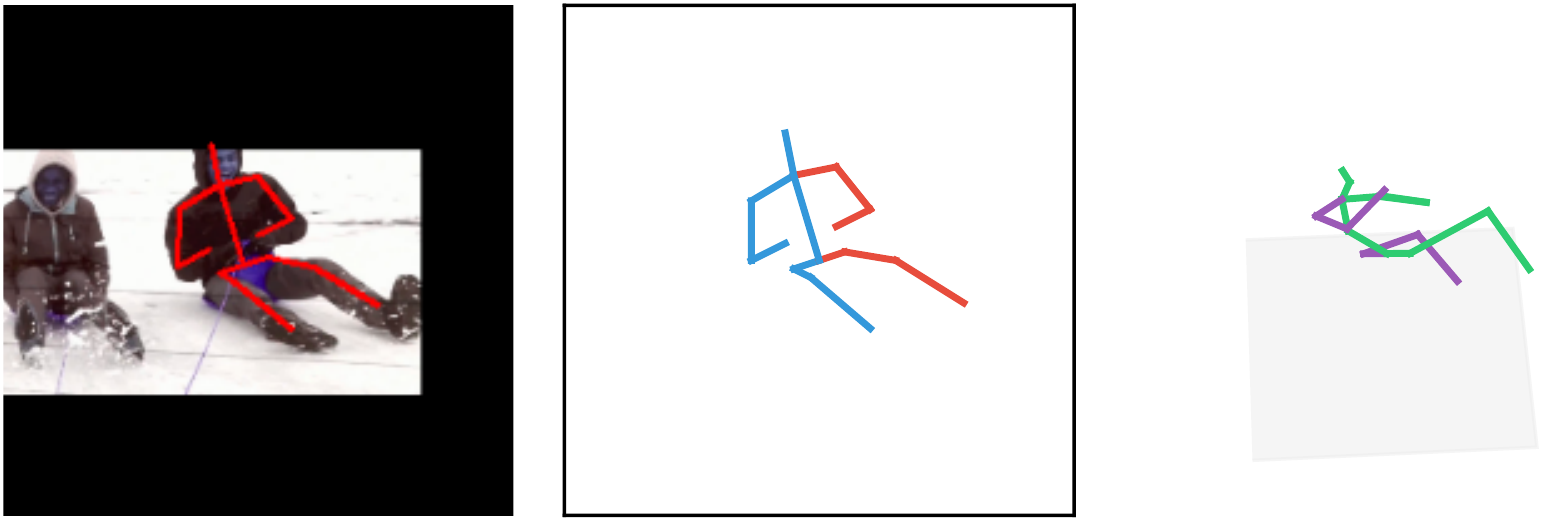} \hfill
  \includegraphics[width=0.29\textwidth]{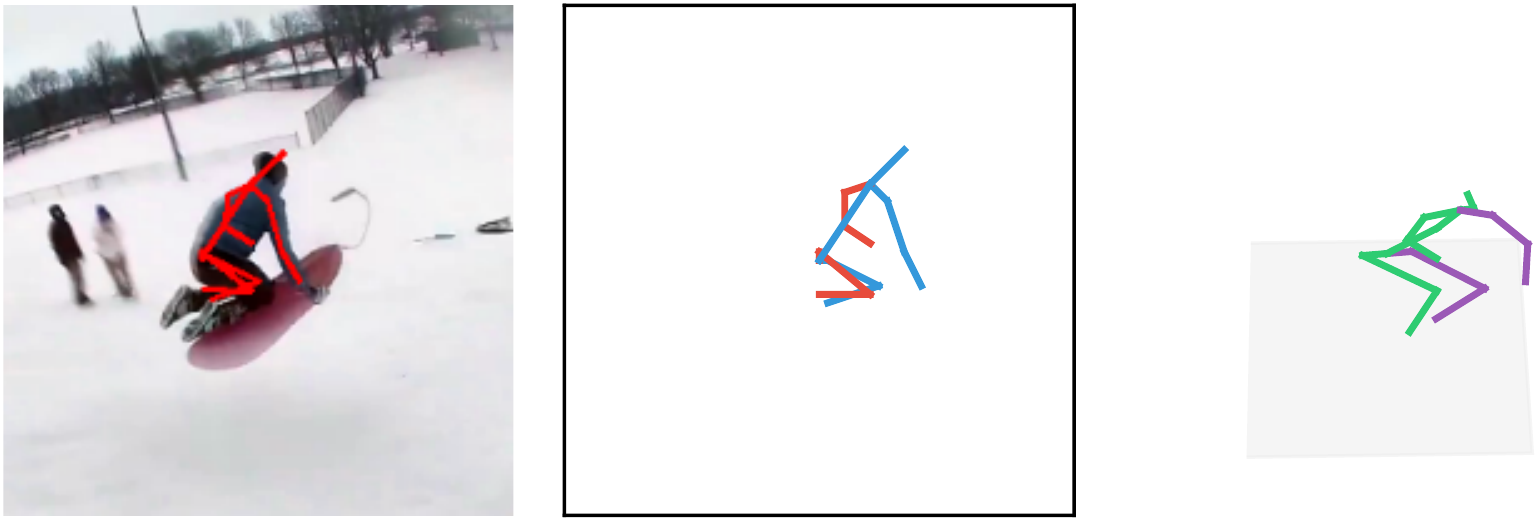}\hfill
\includegraphics[width=0.29\textwidth]{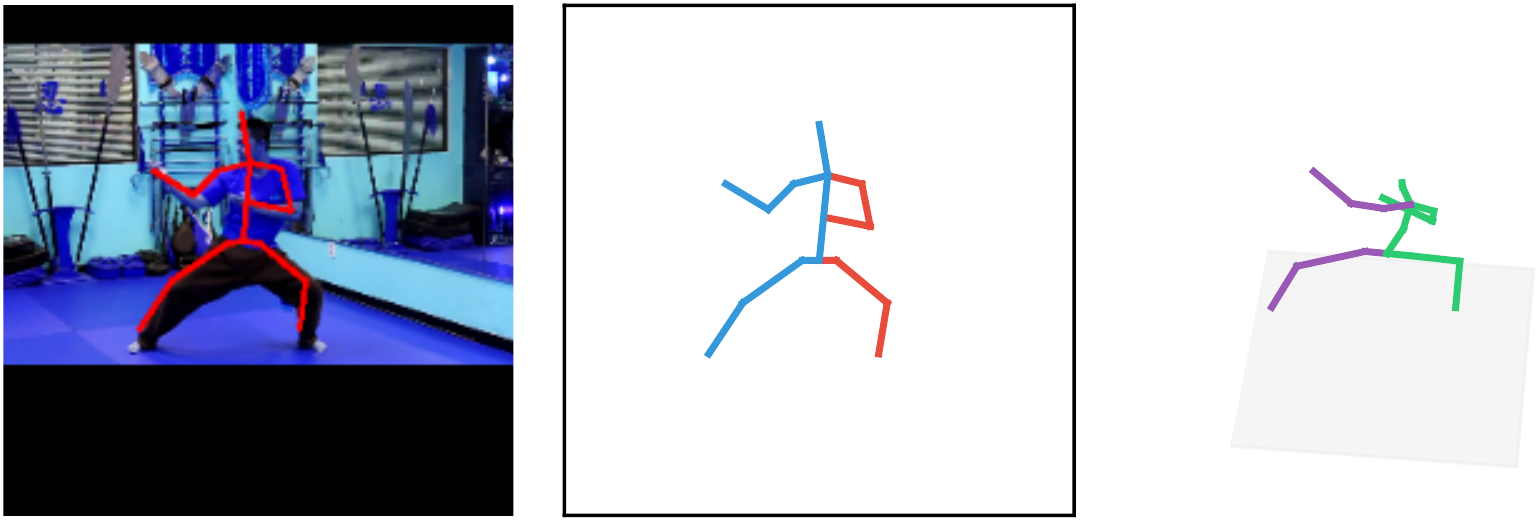}\quad\\

% \quad 
% \includegraphics[width=0.29\textwidth]{latex/images/sample_162-crop.pdf} \hfill
%   \includegraphics[width=0.29\textwidth]{latex/images/sample_268-crop.pdf}\hfill
% \includegraphics[width=0.29\textwidth]{latex/images/sample_311-crop.pdf}\quad\\

\quad 
\includegraphics[width=0.29\textwidth]{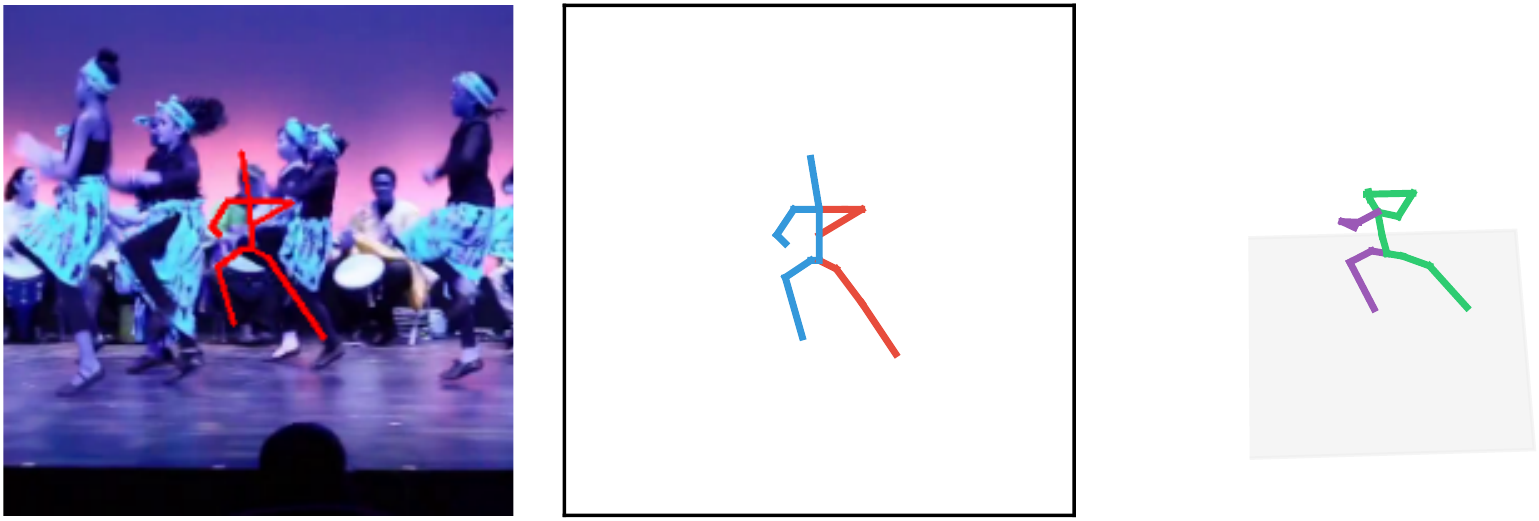} \hfill
  \includegraphics[width=0.29\textwidth]{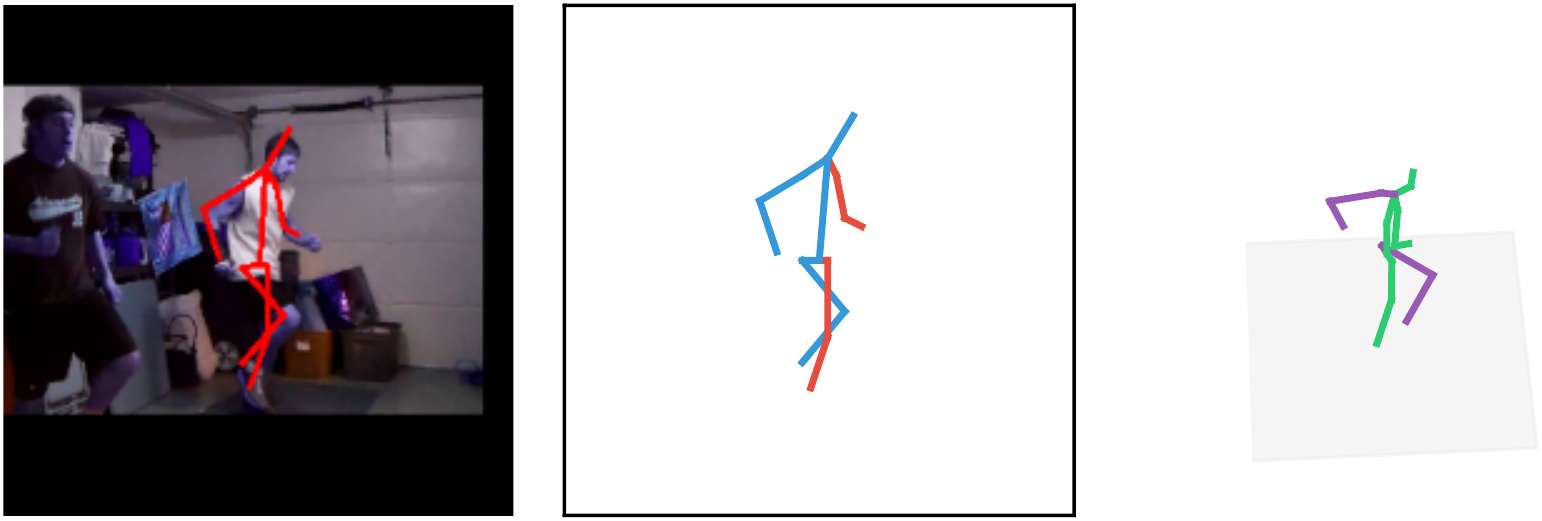}\hfill
\includegraphics[width=0.29\textwidth]{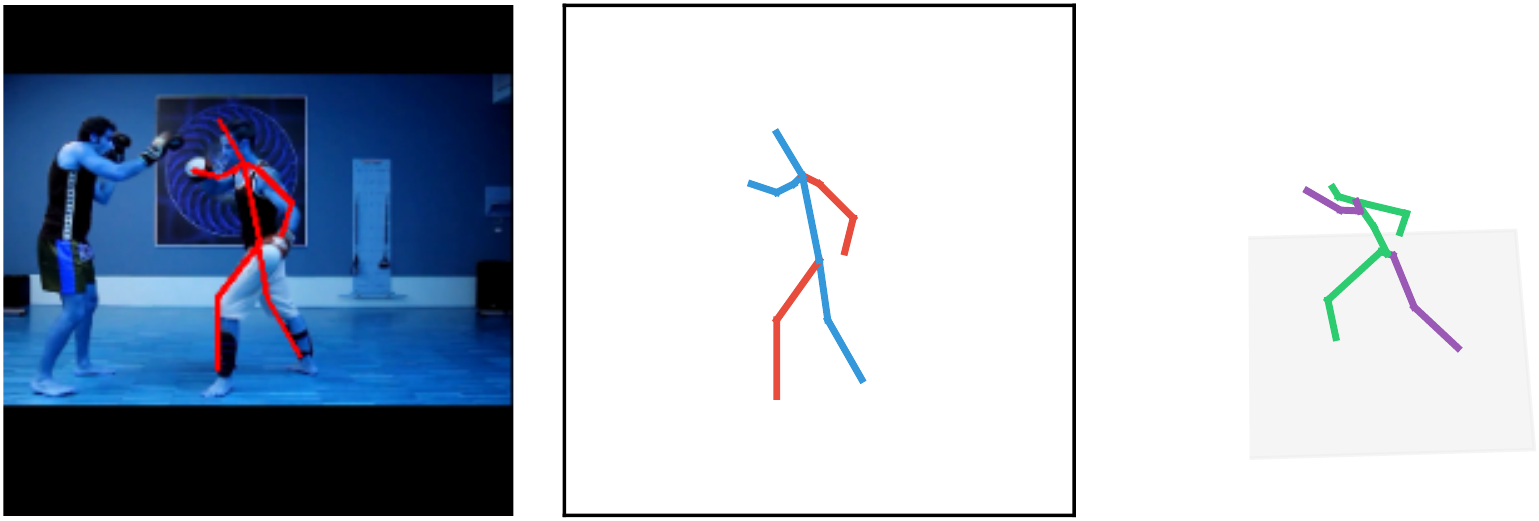}\quad\\

\quad 
\includegraphics[width=0.29\textwidth]{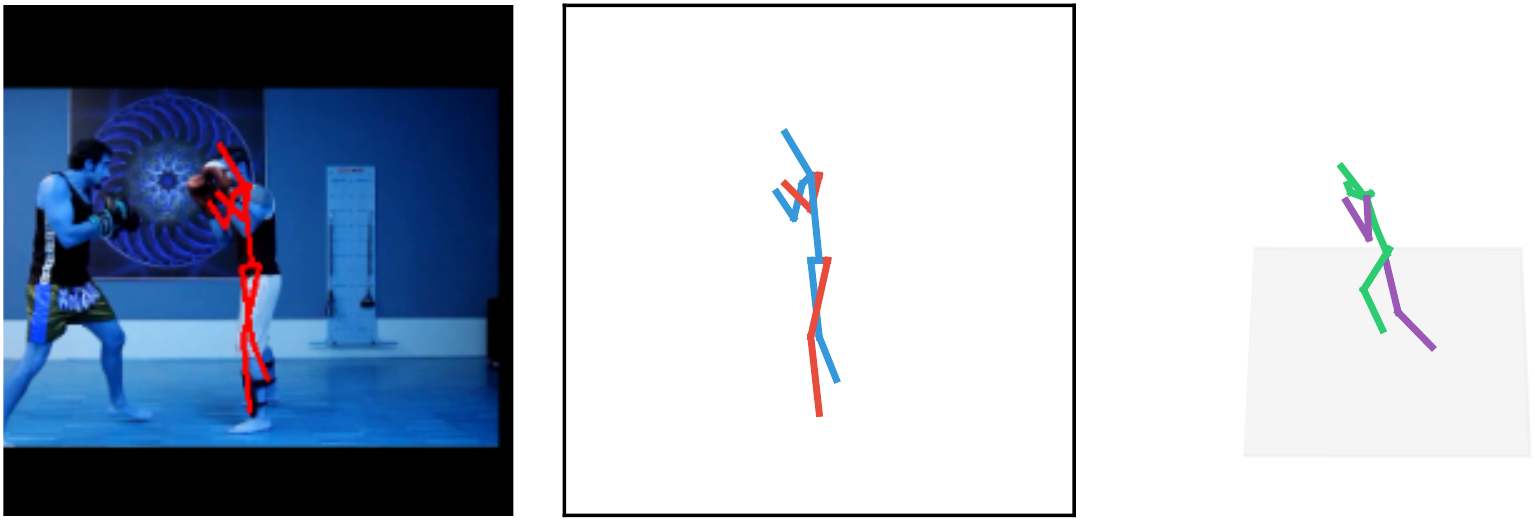} \hfill
  \includegraphics[width=0.29\textwidth]{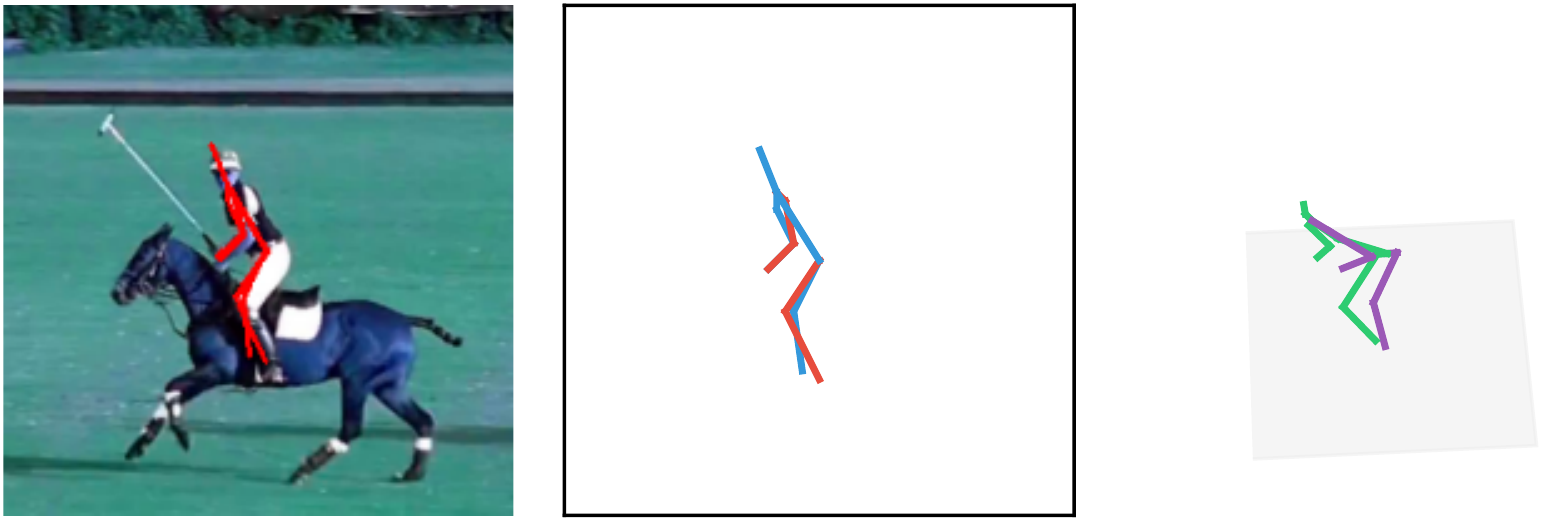}\hfill
\includegraphics[width=0.29\textwidth]{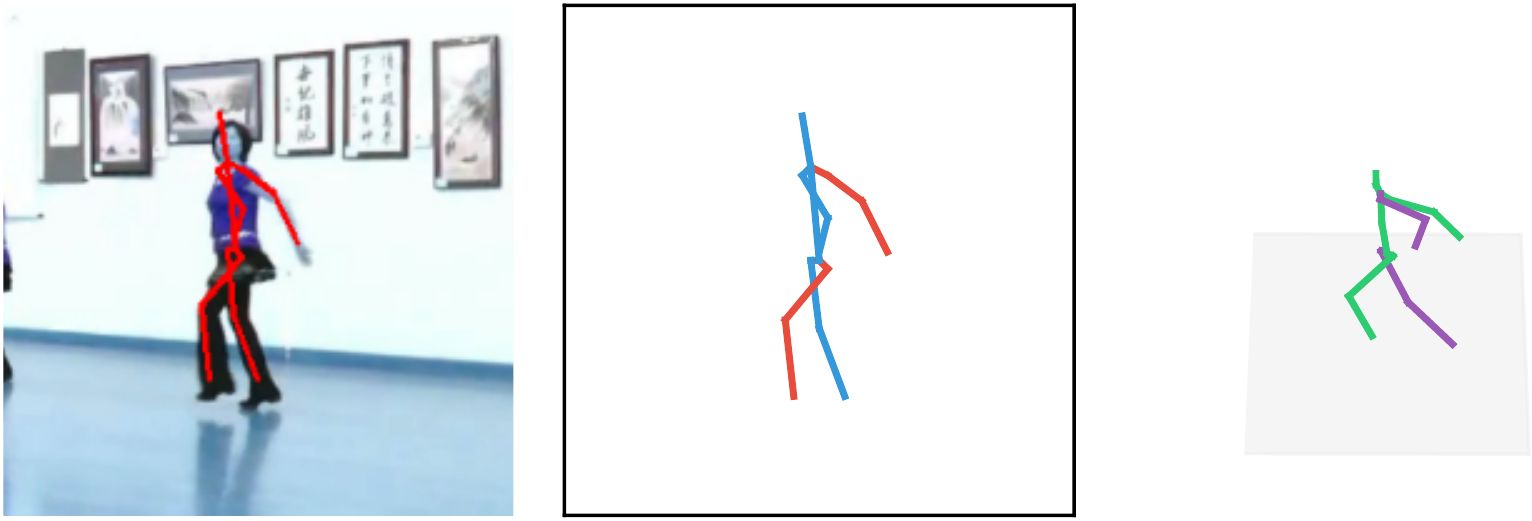}\quad\\

\quad 
\includegraphics[width=0.29\textwidth]{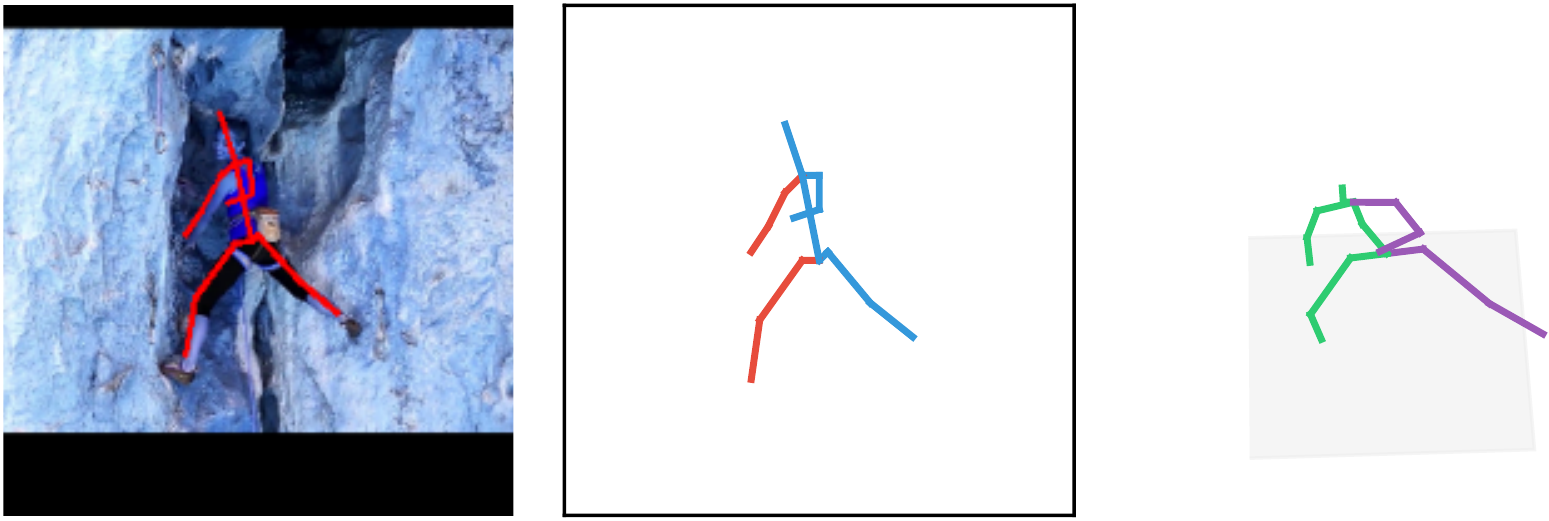} \hfill
  \includegraphics[width=0.29\textwidth]{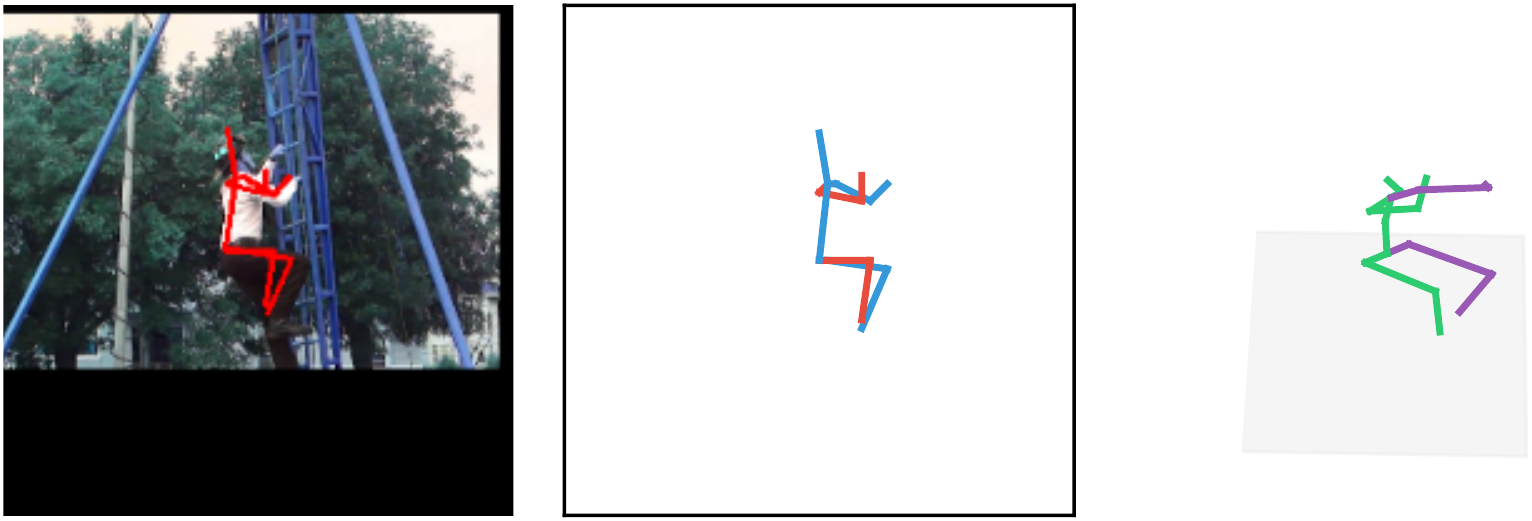}\hfill
\includegraphics[width=0.29\textwidth]{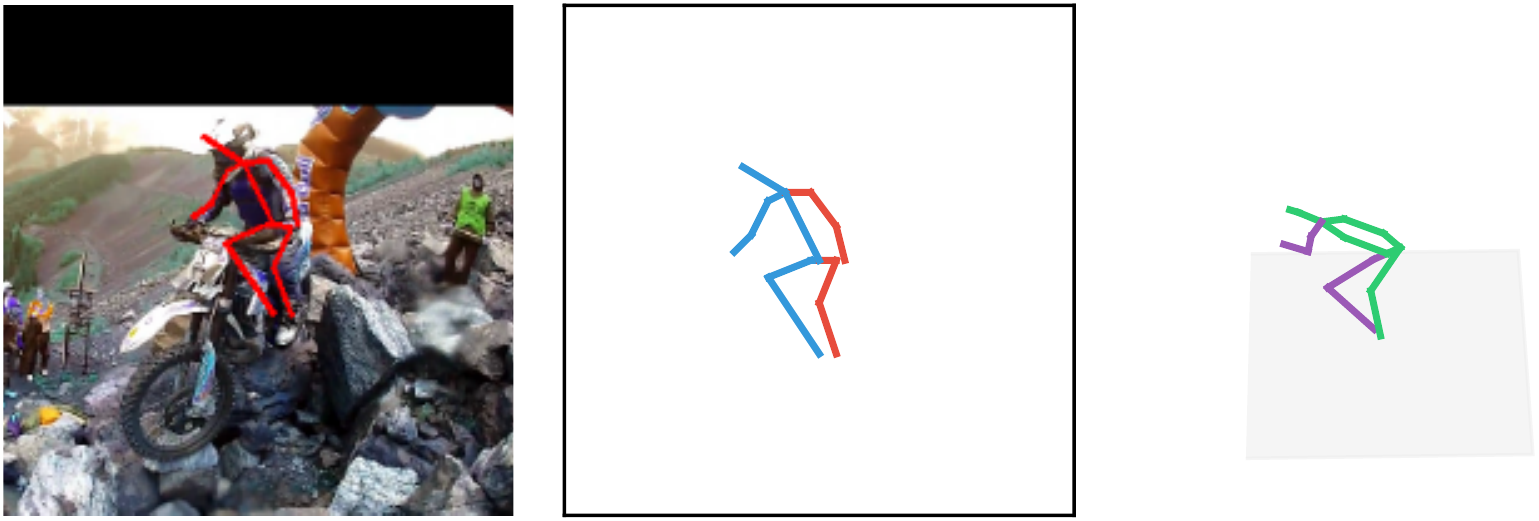}\quad\\

% \quad 
% \includegraphics[width=0.29\textwidth]{latex/images/sample_11-crop.pdf} \hfill
%   \includegraphics[width=0.29\textwidth]{latex/images/sample_638-crop.pdf}\hfill
% \includegraphics[width=0.29\textwidth]{latex/images/sample_10-crop.pdf}\quad\\

  \caption{Qualitative results on the MPII test set. The first and second columns are the input images and output 2D joint detections of the stacked hourglass network, the last column is the 3D pose generated by our network.
  }
  \vspace*{-4mm}
  \label{fig:viz_mpii}
\end{figure*}

\begin{table}[ht]
\caption{Comparison between different number of  kernels}
\vspace{-1mm}
\centering
\small
\setlength{\tabcolsep}{8pt}
\setlength{\textfloatsep}{2pt}
\begin{tabular*}{0.45\textwidth}{ l c c c c } 
 \hline
% &\multicolumn{1}{c}{Methods}\vline
% &\multicolumn{3}{Temporal}{Methods}\vline
% &\multicolumn{2}{c}{Muti-view}\vline \\
 Number of kernels  & 1 & 3 &  5 & 8\\ 
 \hline
  Avg. MPJPE & 62.9 & 55.2 & 52.7  &  \bf 52.6  \\
  \hline
  \vspace{-7mm}
\end{tabular*}
\label{Tab:differentKernelNumber}
\end{table}

\begin{table*}[ht]
\caption{Comparison of our network with and without Dirichlet prior}
\vspace{-1mm}
\centering
\small
\setlength{\tabcolsep}{2pt}
\begin{tabular*}{0.96\textwidth}{ c c c c c c c c c c c c c c c c c } 
 \hline
 &  Direct. & Discuss & Eating & Greet & Phone & Photo & Pose & Purch. &  Sitting & SittingD. & Smoke  & Wait & WalkD. & Walk & WalkT. & Avg.\\ 
 \hline
 
Wo prior & 44.4 & 49.6 & 50.0 &51.0 & \bf 57.3 &63.0 & 46.0 &49.2 &64.1 &78.7 &55.4 &51.4 &56.8 & \bf 43.1 & \bf 44.9 &53.7\\
W prior & \bf 43.8 & \bf 48.6 & \bf 49.1 & \bf 49.8 &57.6 & \bf 61.5 & \bf 45.9 & \bf 48.3 & \bf 62.0 & \bf 73.4 &\bf 54.8 & \bf 50.6 & \bf 56.0 &43.4 &45.5 & \bf 52.7 \\\hline
\vspace{-3mm}

\end{tabular*}
\label{Tab:dirichletPrior}
\end{table*}

\begin{table*}[ht]

\caption{The similarity of the 2D reprojections of all five pose hypotheses  }
\vspace*{-1mm}
\centering
\small
\setlength{\tabcolsep}{2pt}
\begin{tabular*}{0.96\textwidth}{ c c c c c c c c c c c c c c c c c } 
 
 \hline
 &  Direct. & Discuss & Eating & Greet & Phone & Photo  & Pose & Purch. &  Sitting & SittingD. & Smoke & Wait & WalkD. & Walk & WalkT. & Avg.\\ 
 \hline
PCKh@0.5 & 99.6 &  99.5 &  99.6 &  94.9 & 99.5 &  99.7 &  99.9 &  98.8 &  99.0 &  87.6 & 99.6 &  94.6 & 99.1 &  99.2&  99.5 & 98.1 \\ \hline
\vspace{-3mm}
\end{tabular*}

\label{Tab:2D_reprojectionSimilarity}
\end{table*}

\begin{figure*}[ht]
\begin{center}
%\fbox{\rule{0pt}{2in} \rule{.9\linewidth}{0pt}}
\includegraphics[width=0.98\linewidth]{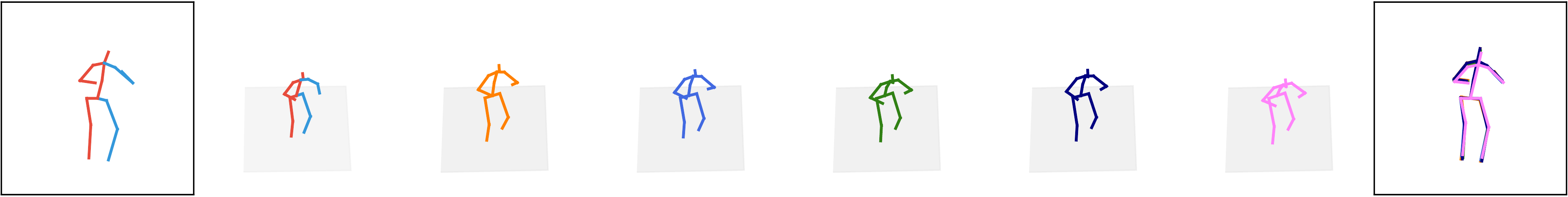}
\includegraphics[width=0.98\linewidth]{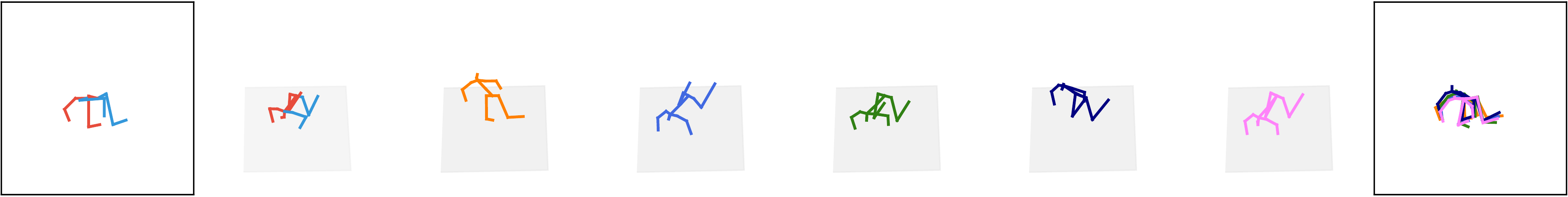}
\includegraphics[width=0.98\linewidth]{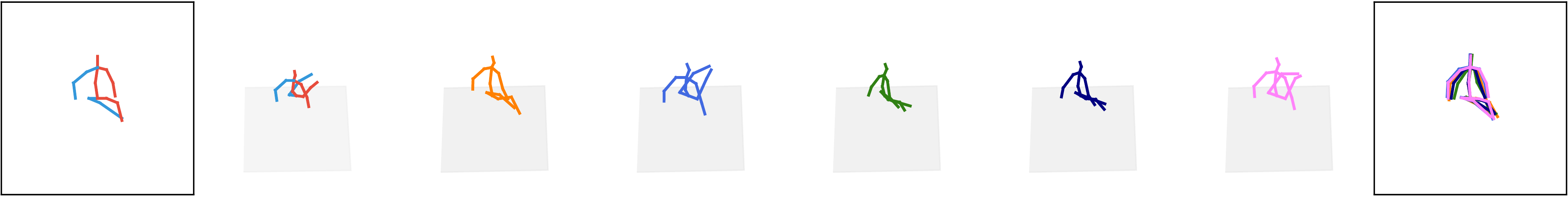}

\end{center}
\vspace{-3mm}
   \caption{3D Pose hypotheses generated by our network. The first column is the input of our network, \ie the 2D joints estimated by the stacked hourglass network. The second column is the ground truth 3D pose, and the third to seventh columns are the hypotheses generated by our network. The last column is the 2D reprojections of all five hypotheses. The corresponding 2D projection and 3D pose are drawn in the same color. (Best view in color) }

\label{fig:hypotheses}
\vspace{-2mm}
\end{figure*}

\subsection{Results on Human3.6M dataset}
We first report our results on the Human3.6 dataset and compare with other state-of-the-art approaches. From the results shown in Table \ref{Tab:MPJPE_Results_Human3.6}, we can see that our method outperforms the others in most cases. Our approach achieves an improvement of  5.5\% compared to the previous best result 55.8 mm \cite{lee2018propagating} and 16.2 \% compared to the baseline architecture \cite{martinez2017simple}. This indicates the efficiency of our approach by generating multiple hypotheses. Moreover, our network outperforms \cite{jahangiri2017generating} which also generates multiple hypotheses by 22.5\%. Following previous work, We show our result under Protocol \#2 \cite{bogo2016keep, moreno20173d} where the estimated pose has been further aligned with the ground truth via a rigid transfermation. The MPJPE error in Table~\ref{Tab:MPJPE_Results_Human3.6} shows that our approach consistently outperforms other approaches. 

It is difficult to disambiguate the multiple 3D pose hypotheses generated by our model in a monocular view because most of them are feasible solutions to the inverse 2D-to-3D problem. Hence, we utilize the multi-view images from the set of calibrated cameras provided by the Human3.6M dataset to disambiguate and verify the correctness of the multiple 3D pose hypotheses generated by our network.
%In order to verify that our model is able to generate the correct pose, we use the camera parameters provided by Human3.6M dataset to help us get the pose from the multiple hypotheses. 
Specifically, we transform the same pose under different cameras into the global world coordinates, and then we choose the pose which is most consistent with the poses from other camera coordinates. Finally, we get our estimated pose by averaging all poses from different camera coordinates. We list our result in Table~\ref{Tab:multi-view} and compare with other state-of-the-art approaches based on multi-view \cite{pavlakos2017harvesting} (spatial constraint) or video (temporal constraint) information~\cite{lee2018propagating,hossain2018exploiting}.
Note that it is however not a fair comparison with other results listed in Table \ref{Tab:MPJPE_Results_Human3.6} because they did not use any multi-view or video information. The results show that our approach has the best performance among both spatial and temporal constraints based methods, indicating the advantage of our approach by generating multiple hypotheses. 

In realistic scenarios, it is common that some joints are occluded and cannot be detected. In order to show that our model can handle with missing joints, we ran experiments with different number of missing joints selected randomly from the limb joints including l/r elbow, l/r wrist, l/r knee, l/r ankle. We show our results in Table~\ref{Tab:missingJoints} and compared with the baseline 2D-to-3D estimator \cite{martinez2017simple} and the GMM based methods \cite{jahangiri2017generating} which also focus on generating multiple hypotheses. The baseline outperforms GMM based methods by a large margin, which indicates the advantage of using deep networks. Moreover, our method improves the baseline for all actions with average error decreased by 10.4mm for both cases, further showing the robustness of our method.

\subsection{Transfer to MPII and MPI-INF-3DHP datasets }

We test our method on the MPII and MPI-INF-3DHP datasets to validate the generalization capacity. Note that we train the feature extractor and hypotheses generator on the Human3.6 dataset which contains data from only the indoor environment. The validation set of MPI-INF-3DHP dataset includes images recorded under three different scenes: 1143 images in studio with green screen background (Studio GS), 1064 images in studio without green screen background (Studio no GS) and 728 images in outdoor environment (Outdoor). We use the 2D joints provided by the dataset as input and compute the 3DPCK. The results in Table~\ref{Tab:resultsMPI-INF-3DHP} show that the 3DPCK of our approach is slightly lower than \cite{mehta2017monocular} even though we did not train on their dataset, indicating the generalization of our network. Moreover, our results for different scenes do not change too much compared to the results of \cite{mehta2017monocular}. This further suggests the domain-invariant capability of the two-stage approach that we adopted. We only give qualitative results for the MPII dataset in Figure~\ref{fig:viz_mpii} since the ground truth 3D pose data is not provided. We can see that our network can be generalized well to outdoor unseen scenes.

\subsection{Ablation Study}
\paragraph{Different number of kernels} Our hypotheses generator is based on MDN where each of the $M$ Gaussian kernels in Eqn.\eqref{eq:prob_density} yields different result. We note that our network cannot fit the data completely if $M$ is too small, while larger $M$ requires more computation resource. We thus train three different models with $M$ setting to 3, 5, 8, respectively. We show the average MPJPE on the Human3.6M dataset in Table~\ref{Tab:differentKernelNumber} and compare them with the baseline method, which is based on single Gaussian distribution. The results suggest that our MDN has better performance than single Gaussion based method. Moreover, the performance does not improve much when $M$ is larger than five. Consequently, we set $M$ to five in our experiments in view of the trade-off between accuracy and computational complexity.
\vspace{-2mm}
\paragraph{Dirichlet prior}
We add a Dirichlet conjugate prior to the distribution of the mixture coefficients $\bm{\alpha} (\textbf{x}$) to prevent overfitting of a single Gaussion kernel to the training data. In order to explore the role of the Dirichlet prior, we compare the performance of our model with and without $\mathcal{L}_{\text{prior}}$. The results are shown in Table \ref{Tab:dirichletPrior}, it can be seen that the performance improves by adding the Dirichlet conjugate prior, especially for the difficult poses in actions ``Sitting'' and ``SittingDown''. The reason is that most of the poses in the Human3.6 dataset are in a standing position, resulting in a worse performance on the  ``Sitting'' and ``SittingDown'' actions. This further indicates that the Dirichlet conjugate prior can prevent overffiting effectively.
\vspace{-2mm}
\paragraph{What is generated by each kernel?}
In order to explore the relation between different hypotheses, we reproject all five pose hypotheses into the image plane and compute the difference between projections and the 2D input joints. We adopt the PCKh@0.5 score \cite{newell2016stacked} which is the standard metric for 2D pose estimation to measure the difference. The high PCKh@0.5 score in Table~\ref{Tab:2D_reprojectionSimilarity} suggests that all the five hypotheses have almost the same 2D reprojections which are consistent with the 2D input. Note that we do not add any constraint as \cite{jahangiri2017generating} did to force all hypotheses to be consistent in the 2D reprojections.  

We give several visualization results in Figure~\ref{fig:hypotheses} to further illustrate the relations between all pose hypotheses. 
As described by Eqn.~\eqref{eq:prob_density_single},
each Gaussian kernel can be seen to generate the same hypotheses
for simple pose with less ambiguity, \eg standing (first row).
This means that single Gaussian distribution is sufficient for simple poses. In comparison, our network can be seen to generate different hypotheses for challenging poses like ``GettingDown'' or ``SittingDown'' (second and third rows) due to two reasons. Firstly, our network receives lesser information on this type of poses since most of the poses in the Human3.6 dataset are the ``standing" poses. Secondly, there are more ambiguities and occlusions for the ``GettingDown'' or ``SittingDown'' poses. As a result, our network generates multiple pose hypotheses to mitigate the increase of the uncertainty.
We also visualize the 2D reprojections of all hypotheses in the last column. We indicate the corresponding 2D reprojection and 3D pose with the same color. 
The overlaps between the 2D reprojections further validate that our network generates hypotheses that are consistent in the 2D image coordinates.

\section{Conclusion}

In this work, we introduce the use of a mixture density network to generate multiple feasible hypotheses for the inverse problem of 3D human pose estimation from 2D inputs. Experimental results show that our network achieves state-of-the-art results in both best hypothesis and multi-view settings. Furthermore, the 3D pose hypotheses generated by our network are consistent in the 2D reprojections suggests that the hypotheses model the ambiguity along the depth of the joints.
Results on the MPII and MPI-INF-3DHP datasets further show the generalization capacity of our network.

{\small
  \bibliographystyle{ieee}
  \bibliography{egbib}
}
%In this work, we explored the idea of generating multiple 3D hypotheses to alleviate the ambiguity problem. We combined conventional 2D-to-3D human pose estimator with the MDN to generate multiple hypotheses. Our network performs better than existing 3D human pose estimation methods especially for challenging poses. Moreover, The pose hypotheses generated by our network are consistent in 2D projections indicates that our approach is trying to solve the inverse 2D-to-3D problem by generating multiple hypotheses. Results on MPII and MPI-INF-3DHP datasets further show the generalization capacity of our network.

\end{document}